# Hebrew letters Detection and Cuneiform tablets Classification by using the yolov8 computer vision model

Elaf A. Saeed [1*], Ammar D. Jasim [2], and Munther A. Abdul Malik [3]

Department of System Engineering, Collage of Information Engineering, AL-Nahrain University, Baghdad, Iraq [1,2]

Department of History, College of Literature, Baghdad University, Baghdad, Iraq [3]

**Abstract**

*Cuneiform writing, an old art style, allows us to see into the past. Aside from Egyptian hieroglyphs, the cuneiform script is one of the oldest writing systems. It emerged in the second half of the fourth millennium BC. The majority of people believe that it was originally created by the Sumerians in southern Mesopotamia. Many historians place Hebrew's origins in antiquity. For example, we used the same approach to decipher the cuneiform languages; after learning how to decipher one old language, we would visit an archaeologist to learn how to decipher any other ancient language. To expedite this procedure, we propose a deep-learning-based sign detector method to identify and group cuneiform tablet images according to Hebrew letter content. The Hebrew alphabet is notoriously difficult and costly to gather the training data needed for deep learning, which entails enclosing Hebrew characters in boxes. We solve this problem by using pre-existing transliterations and a sign-by-sign representation of the tablet's content in Latin characters. We recommend one of the supervised approaches because these do not include sign localization: We Find the transliteration signs in the tablet photographs by comparing them to their corresponding transliterations. Then, retrain the sign detector using these localized signs instead of utilizing annotations. Afterward, a more effective sign detector enhances the alignment quality. Consequently, this research aims to use the Yolov8 object identification pretraining model to identify Hebrew characters and categorize the cuneiform tablets. Images illustrating Hebrew passages have been culled from a Hebrew-language book. This book is known as the Old Testament, and it was organized into around 500 illustrations to aid in reading and pronouncing the characters. Another ancient document was recently discovered in Iraq, dating back to 500 BC. It reached over a thousand photos after pre-processing and augmentation. The CDLI website and the Iraqi Museum have compiled photographs of cuneiform tablets, with over a thousand photos available in each language. With a recall of 89.8 percent, a precision of 93.2 percent, and a mean Average Percentage of 50 (mAP50) of 92%, 22 letters were successfully detected. The classification's accuracy is top1 96% and top5 100%.*

## 1. Introduction

Many applications rely on real-time object identification, spanning several domains, including augmented reality, video surveillance, autonomous cars, and robotics. Because of its remarkable speed and accuracy, the YOLO (You Only Look Once) framework has become a prominent object identification technique. It enables reliable and fast object recognition in images. Since its inception, the YOLO family has evolved through several revisions, with each iteration enhancing the previous one to circumvent limitations and boost performance [1].

The writing system originated from symbols representing concepts derived from the Sumerian language. The Assyrian and Babylonian languages emerged due to the progressive evolution of symbolic writing via many developmental phases [2]. In contrast to the hieroglyphic visual language, the cuneiform system is characterized by its greater emphasis on verbal expression and its use of particular vocabulary to convey precise meanings.

Over 10,000 cuneiform tablets have been found at the International Museum and the Iraqi Museum, the latter having over 2,000 of these tablets [3]. The cuneiform script in Mesopotamia evolved into the Assyrian cuneiform language. The direction of writing shifted to right-to-left, and symbols were inscribed into stone or clay tablets. The cuneiform alphabet comprises around 600 letters, each composed of one or more symbols [4].

Hebrew is a non-European language with a non-alphabetic writing system; this developmental research looks into the aforementioned consequences in Hebrew. As with Arabic, Hebrew is a Semitic language that uses the abjad writing system. Like Arabic, Hebrew exclusively represents consonants and is read from right to left. The Hebrew alphabet consists entirely of 22 consonants. There are two ways vowels are represented: (i) with four letters that may be used as both vowels and consonants and (ii) with thirteen vowel signs that can be used, like diacritics. Most Hebrew literature aimed at beginners, poetry, children's books, and religious materials employ these vowel signs [2].

Our goal is to help scientists understand Hebrew letters so they may better conduct their studies. Our primary objective is to develop a Hebrew letter detector capable of precisely identifying and locating the sign's pronunciation within a bounding box. Additionally, it creates a paradigm for categorizing cuneiform tablets. The scripts were categorized as Assyrian, Cuneiform, or Babylonian [4].

This study uses a character-based strategy to identify Hebrew letters by their bounding boxes. Furthermore, a pretraining model labels the Hebrew letter's sound. Scientists rely on bounding boxes at the letter level to help them understand how the detector makes decisions [4]. This study's originality lies in its exhaustive data collection, which includes more than 400 images of Hebrew characters. To train the model for accurate identification, we used state-of-the-art algorithms like YOLOv8 [5]. As a novel pre-training model for Hebrew character recognition, the paper presents YOLOV8, the most recent YOLO object identification system version. Not only were the Hebrew letters correctly detected, but significant outcomes were also achieved.

## 2. Literature Survey

[6] Using the YOLOv8 object identification pretraining model, this effort aimed to detect signs on Assyrian cuneiform tablets. We improved and pre-processed approximately eight hundred photographs of Assyrian tablets obtained from the Iraq Museum till their size exceeded two thousand. Consequently, eleven more Assyrian references were located, with an accuracy rate of 71%, a recall of 85%, and a mean average precision (mAP) of 82% at 50 epochs. By making it easier to recognize cuneiform signs and pick and pronounce the present Assyrian dialect, this work enabled researchers to read with a pre-trained model.

[7] This research delves into the challenges faced by the character recognition (CR) system when evaluating the Great Isaiah Scroll pictures. We created a new dataset using images

of individual letters taken from the scroll. Additionally, four Convolutional Neural Network (CNN) models were successfully tested on our dataset in this research. Among the convolutional neural network (CNN) models tested, AlexNet and LeNet-5 performed the best when correctly detecting ancient Hebrew letters written by hand. With an impressive 94% test accuracy, these models display consistently low loss rates and accuracy fluctuations.

In [8], a new R-CNN architecture approach is proposed to classify and localize pegs. Three-dimensional models of 1977 cuneiform tablets from Frau Professor Hilprecht's collection, available as open data, were used. The approach consists of a pipeline of two components: a signal detector and a wedge detector. The signal detector uses a RepPoints model with a ResNet18 backbone to locate individual cuneiform characters in a tablet section image. The wedge detector is based on the Point RCNN approach.

Given in [9], Cuneiform writing, which has been used for over three millennia and at least eight main languages, is a 3D script imprinted into clay tablets. Digital tools for processing this script were developed. Approximately 500 annotated tablets comprise the HeiCuBeDa and MaiCuBeDa databases, which were developed and utilized. I developed a new mapping tool to help people annotate 3D models and photos using an innovative OCR-like method for mixed-image data.

The strategy in [10] uses SIFT-Descriptors in a bag-of-features (BoF) way. This study aims to offer segmentation-free cuneiform sign identification by combining a patch-based (sliding window) technique with hidden Markov models (HMMs).

The [11] basic elements, namely the strokes of the cuneiform characters shown in photographs of old cuneiform tablets, are the focus of this research. Using modern computer vision techniques, we want to make optical character recognition (OCR) more efficient. Using two-dimensional images instead of three-dimensional models is a key differentiator between our method and earlier approaches. The profusion of freely available online archives housing a substantially larger quantity of 2D photographs is the driving force behind this decision. The purpose of creating this program was to make it possible to employ convolutional image filtering techniques to emphasize stroke letters. To make more of an object's edges and less of its backdrop, these edge filters are a common tool.

### 3. Yolov8 Model

A number of tasks, including instance segmentation, object recognition, and picture classification, are within the capabilities of the most recent version of the YOLO model, which is referred to as YOLOv8 [12].

The cross-stage partial bottleneck with two convolutions (C2f) module was used in place of the cross-stage partial layer (CSPLayer) to achieve the goal of making the backbone of YOLOv8 comparable to that of YOLOv5 [13]. Within the framework of YOLOv8, an anchor-free model that incorporates a decoupled head is responsible for independently handling objectless, classification, and regression tasks. The softmax function illustrates the object probabilities that are included inside each class. Binary cross-entropy is used for classification loss in the YOLOv8 method, distribution focal loss (DFL) [14] is used for bounding box loss, and complete intersection over union (CIoU) is used for complete intersection over union [15].

## 4. The Proposed Model

*Figure 1* provides a comprehensive overview of the models that have been presented.
Once the Roboflow platform has labeled all the classes in each image, the initial step in the detection model process is to input the image into the model.
Subsequently, the preprocessing and augmentation approach will be employed to enhance and enlarge the dataset substantially.
During the training phase, the Yolov8 model was exposed to the dataset for integration. After finishing the training, I initiated the model testing procedure by inputting images from the testing dataset. The results of this procedure involved the formation of a dataset comprising all the Hebrew letters, which were subsequently categorized based on the sound of each letter. The Google text-to-speech (gtts) [16] application programming interface (API) was used to convert each label from text to speech.
The cuneiform tablets are categorized using the same methods, although, unlike other models, labeling is not mandatory in this approach. *Figure 2* shows the Proposed Models Flowchart for Detection and Classification.

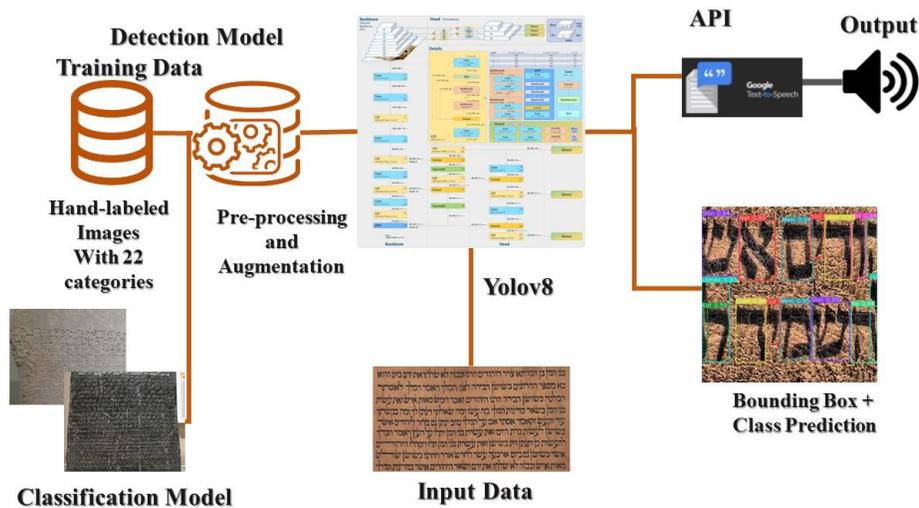

**Figure 1** The Proposed Model Block Diagram

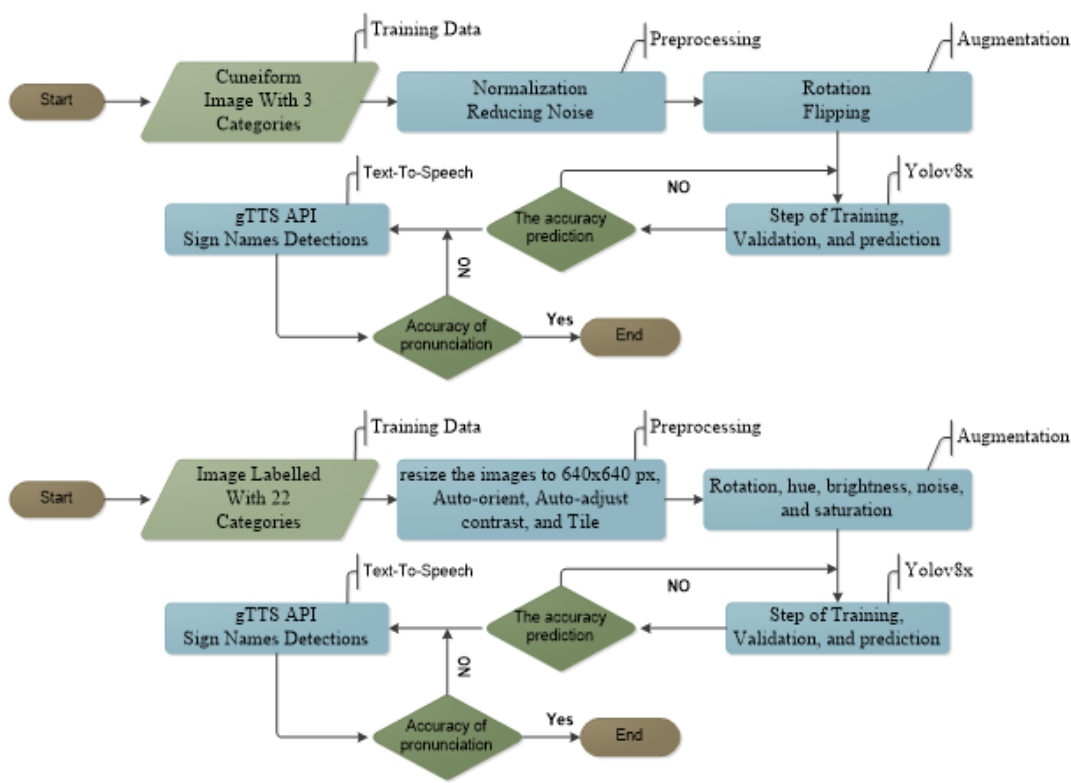

**Figure 2** The Proposed Models Flowchart for Detection and Classification

## 3.1. Dataset Creation
### 3.1.1. Dataset Collection
Train high-quality models with a large number of carefully annotated input pictures. Roughly 1500 pictures are suggested for accurate class detection [17]. An assemblage of Hebrew inscriptions has been gathered. The collected data set comprises images from the Old Testament book depicting the genesis of creation and images of a recently discovered Hebrew manuscript in Iraq. Furthermore, a collection of deceased photographs was posted on the internet. The data collection process faced challenges due to the presence of Hebrew texts with overlapping and diminutive letters, making their identification somewhat arduous. Additionally, identifying Hebrew letters was limited to 22, necessitating a diverse range of visual data and a substantial quantity to enhance the training. There are two options to acquire a greater quantity of labeled data: either increase the size of the dataset or employ data augmentation techniques to amplify the dataset's size. Expanding the dataset can improve the model's capacity to detect things precisely. *Figure 3* displays several images of Hebrew texts.

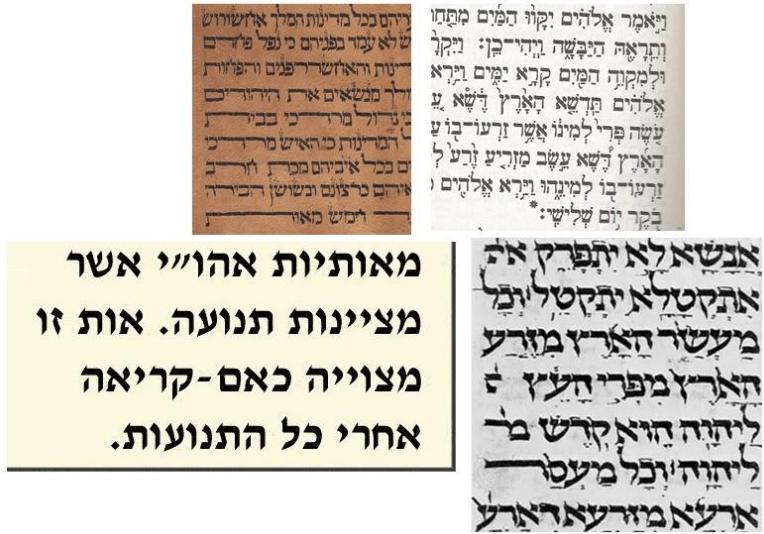

**Figure 4** Some pictures of Hebrew texts

The Iraq Museum provided images of fourteen stone tablets with Assyrian inscriptions for the categorization model. At the same time, the Cuneiform Digital Library Initiative (CDLI) website was used for the collection of all other images. Regarding the Sumerian and Babylonian texts, some were culled from the CDLI website, while others are variations on existing images found online. There were 1327 Assyrian pictures, 1435 Sumerian images, and 1321 Babylonian images. Figure 5 displays the pictures from the dataset of cuneiform tablets.

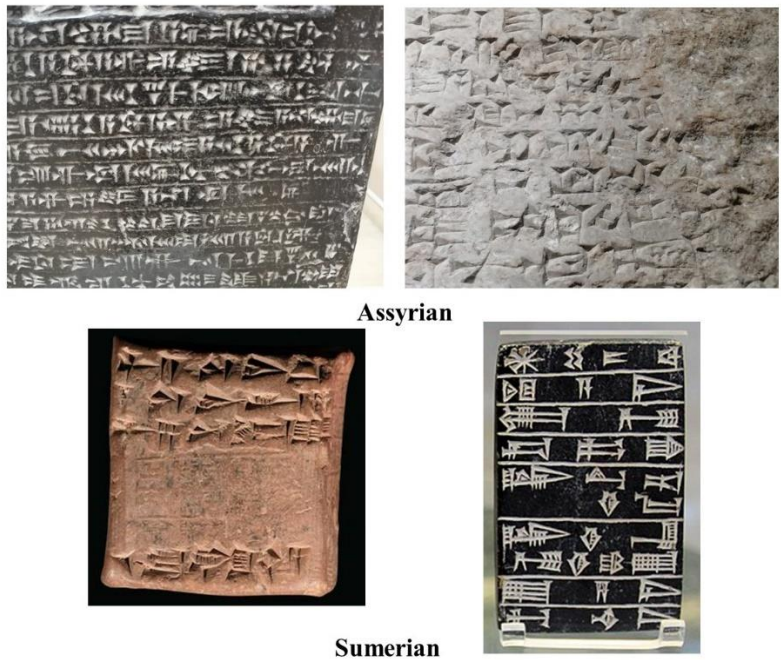

**Figure 5** Some Pictures of the Classification Dataset

### 3.1.2. Dataset Labelling

All 22 Hebrew letters were identified, as shown in Table 1, using the roboflow annotation tool, one of the tools used in labeling. The largest number for annotation is 2381 for the letter Vav, and the smallest is 123 for the letter Tet, as shown in *Figure 6* It shows the number of annotations that were reached after doing the labeling. Verbal signs are printed in another language to make the signs easier to read. For example, the tag (א) is pronounced as Alef, and the remaining tags (ב) are pronounced as Bet.

The database was compiled, and the detection was confirmed to be correct with the help of ***Prof. Bahaa Amer***, a specialist in Hebrew writing at the University of Baghdad, College of Literature.

**Table 1** Dataset Split

| L. No | Hebrew Letters | Letters Name |
|---|---|---|
| 1 | א | Alef |
| 2 | ב | Bet |
| 3 | ג | Gimel |
| 4 | ד | Dalet |
| 5 | ה | He |
| 6 | ו | Vav |
| 7 | ז | Zayin |
| 8 | ח | Chet |
| 9 | ט | Tet |
| 10 | י | Yod |
| 11 | ך, כ | Kaf |
| 12 | ל | Lamed |
| 13 | ם, מ | Mem |
| 14 | ן, נ | Nun |
| 15 | ס | Samech |
| 16 | ע | Ayin |

| 17 | ף , פ | Feh |
| 18 | ץ, צ | Tsadeh |
| 19 | ק | Qof |
| 20 | ר | Resh |
| 21 | ש | Shin |
| 22 | ת | Tav |

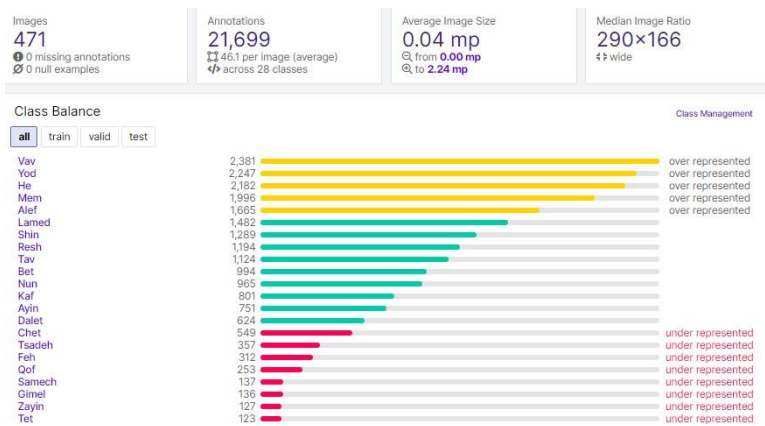

**Figure 6** Number of annotations to each letter

The classification model does not need labeling using the Roboflow platform; it is just the dataset that was split to train, validate, and test images after the pre-processing and augmentation for the dataset and then entered into the model.

### 3.1.3. Enhance the dataset

Roboflow was used to annotate, pre-process, and enrich the dataset to enhance the dataset. With a total of 21,699 annotations, the collection includes 471 tagged photos. The auto-orient, auto-adjust constraints and the resize image (640 × 640 pixels) were utilized. After that, you may adjust the brightness, hue, saturation, and rotation between -15° and +15°, -15% and +15%, and -31% and +31%, respectively, and noise up to 1.8% of pixel augmentation.

When used in pre-processing, auto-orient changes the orientation of the picture's pixels, which in turn changes the orientation of the objects; this aids target detection in cases when the image is rotated. Implementing automated limitations might enhance our neural networks' capacity to understand the objects' nature. Edges are made more distinct with contrast pre-processing because it increases the contrasts between nearby pixels. The permitted input picture size for yolov8 is 640x640px, therefore you'll need to resize it to that size.

Data augmentation is effective because it enhances the semantic breadth of a dataset. A widely used data augmentation technique involves applying random rotations to the data. The source image undergoes a random rotation in either a clockwise or anticlockwise direction. In object detection problems, updating the bounding box to include the generated object is imperative. The brightness Introduces fluctuations in image luminosity to enhance the adaptability of your model to variations in lighting conditions and camera configurations. To encourage a model to investigate various color schemes for objects and scenes in the input photos, hue augmentation randomly alters the color channels of the input images. This technique is beneficial for verifying that a model is not simply recalling the colors of a certain object or scene. Similar to hue modification, saturation augmentation alters the image's vibrancy. Grayscale results from a fully desaturated picture, whereas muted colors are shown in a partially desaturated one. On the other hand, increasing the saturation of an image intensifies the colors, shifting them closer to the primary hues. Noise is a type of flaw that machines find particularly vexing compared to human comprehension. While humans can disregard noise or incorporate it into the appropriate context effortlessly, algorithms encounter difficulty in doing so. The phenomenon referred to as adversarial attacks originates from manipulating pixels in a manner that is invisible to humans yet significantly impacts a neural network's capacity to anticipate outcomes accurately. *Figure 7* shows the images accompanied by labels used for training and validation purposes. The pictures show how it looks after preprocessing and augmentation.

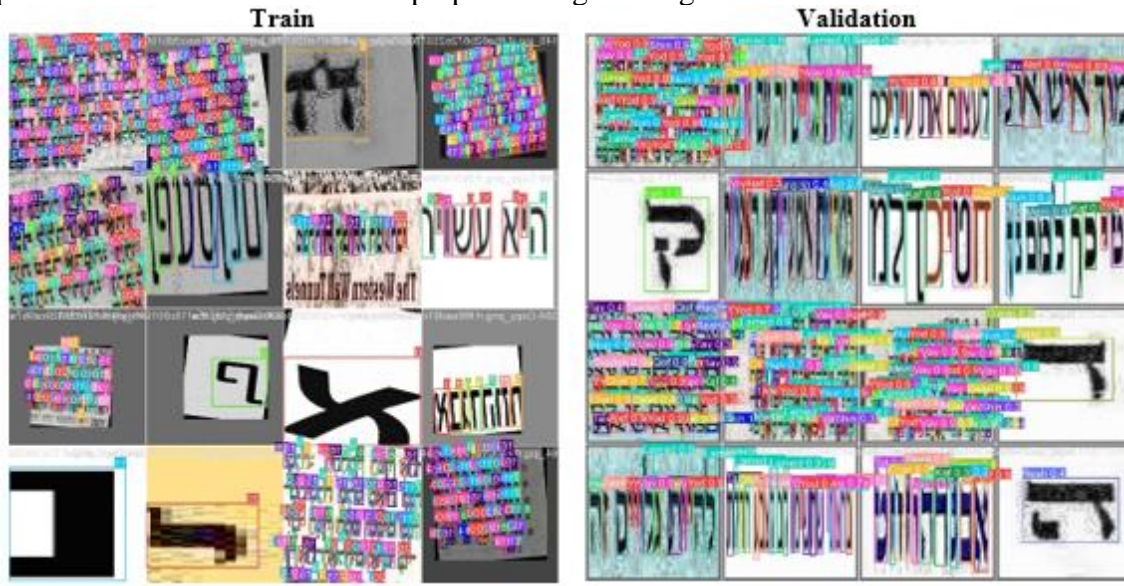

**Figure 7** Images Accompanied with Labels Used for Training and Validation Purposes

Table 2 shows the results of the two variant versions (V) of dataset splitting that were used to determine the training, validation, and testing ratios.

**Table 2** Hebrew Dataset Split

| Version No. | Train – Val - Test | # of all images | Pre-processing | Augmentation | Background images |
|---|---|---|---|---|---|
| 1 | 993 - 94 – 46 | 1133 | 1- Auto-orient. | 1- Rotation. | Without Background images |

| 2 | | | 2- Resize (640x640 px) 3- Auto-adjust contrast. | 2- Brightness. 3- Hue. 4- Noise. | With Background images |

Different numbers of images were chosen for training and testing. The initial version utilized three pre-processing techniques (auto-orient, resize, auto-adjust) and four augmentation techniques (rotation, brightness, hue, noise), resulting in the generation of 1133 images (train=993 - 88%, val=94 - 8%, test=46 - 4%) using the roboflow tool. In the second version, the training was made with a background image to enhance accuracy and reduce FP.

The classification model for pre-processing, Histogram equalization, was employed to normalize the photos by equalizing light distribution. As part of the pre-processing stage, a bilateral filter was employed to reduce noise. This filter effectively smooths the pixels and enhances their edges. The applied augmentation involves rotating within a range of -90 to +90 degrees, which enhances the model's adaptability to photos captured from various perspectives. The second augmentation approach involves vertical and horizontal flips on the images. These two actions were undertaken to enhance the precision of the model, as shown in *Table 3*.

**Table 3** Classification Dataset Split

| | Total Number of Images | Train – Val - Test | Pre-processing | Augmentation |
|---|---|---|---|---|
| Classification | Sumerian = 1440 Assyrian = 1330 Babylonian= 1330 | 70% - 15% - 15% | - Normalization - Noise reduction | - Rotation - Flipping |

### 3.1.4. Performance Evaluation

The detection model achieved impressive assessment results upon training the model using a customized dataset. Specifically, the mean average precision at 50% intersection over union (mAP50) was 92%, the mAP50-90 was 73.5%, the precision was 93.2%, and the recall was 89.8% for the initial version after 100 epochs. The *table (4)* displays the performance evaluation.

**Table 4** Performance Evaluation

| V | Epochs | mAP50 | mAP50-90 | Precision | Recall |
|---|---|---|---|---|---|
| 1 | 100 | 92% | 73.5% | 93.2% | 89.8% |
| 2 | 100 | 91.3% | 72.5% | 92.6% | 88.9% |

The confusion matrix is a method for assessing the performance of a detection model. The y-axis represents the predicted class, whereas the x-axis represents the actual class [29].

Figure (8) demonstrates that classes with more labels have more erroneous predictions than classes with fewer labels. As previously stated, the Tet=123 labels and the vav=2,381 labels.

The maximum true positive (TP) value in versions V1 and V2 was 0.99 for the Ayin class, while the lowest was 0.62 for the Gimal class.

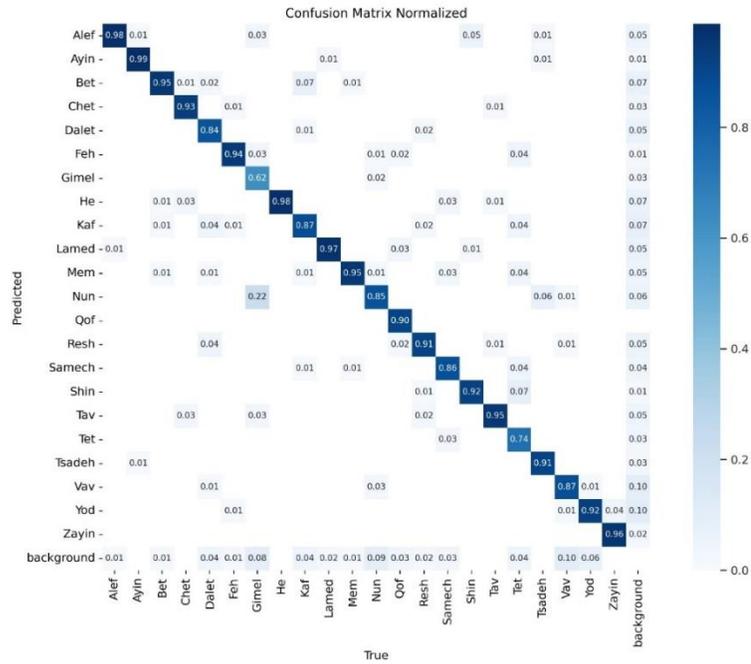

*Version 1*

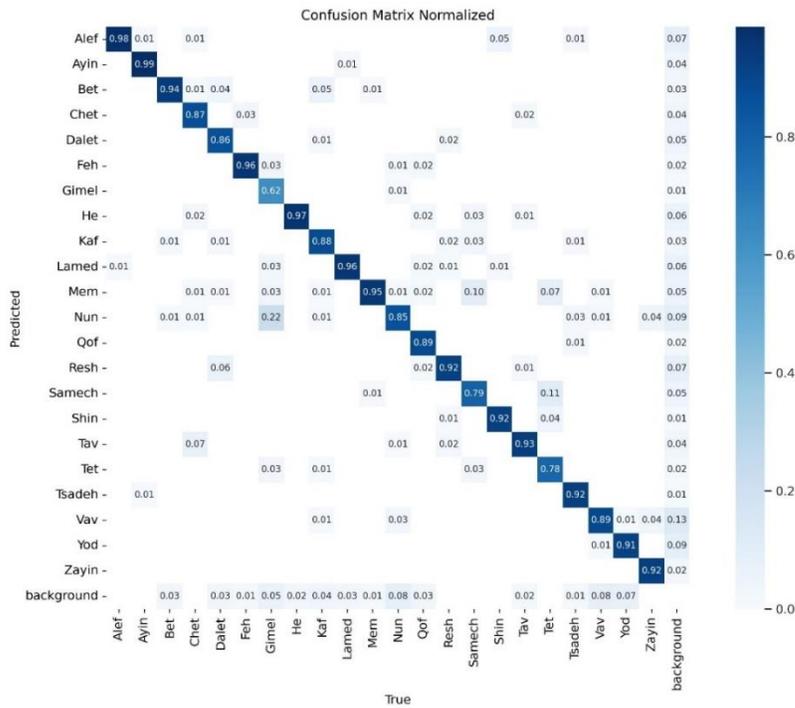

*Version 2*

**Figure 8** Dataset separated into three versions (V) and their respective confusion matrices

Figure (9) displays the plots of F1-confidence, precision-recall, recall-confidence, and precision-confidence for various IOU levels.

The F1-confidence curve illustrates the balance between false positive and false negative predictions. The precision-confidence curve indicates that the average outcome for all categories is 0.99 at a confidence level of 1.00. In the precision-recall curve, the mAP50 value is 0.92. Finally, the mean result of the recall-confidence curve for all classes at a threshold of zero is 0.94.

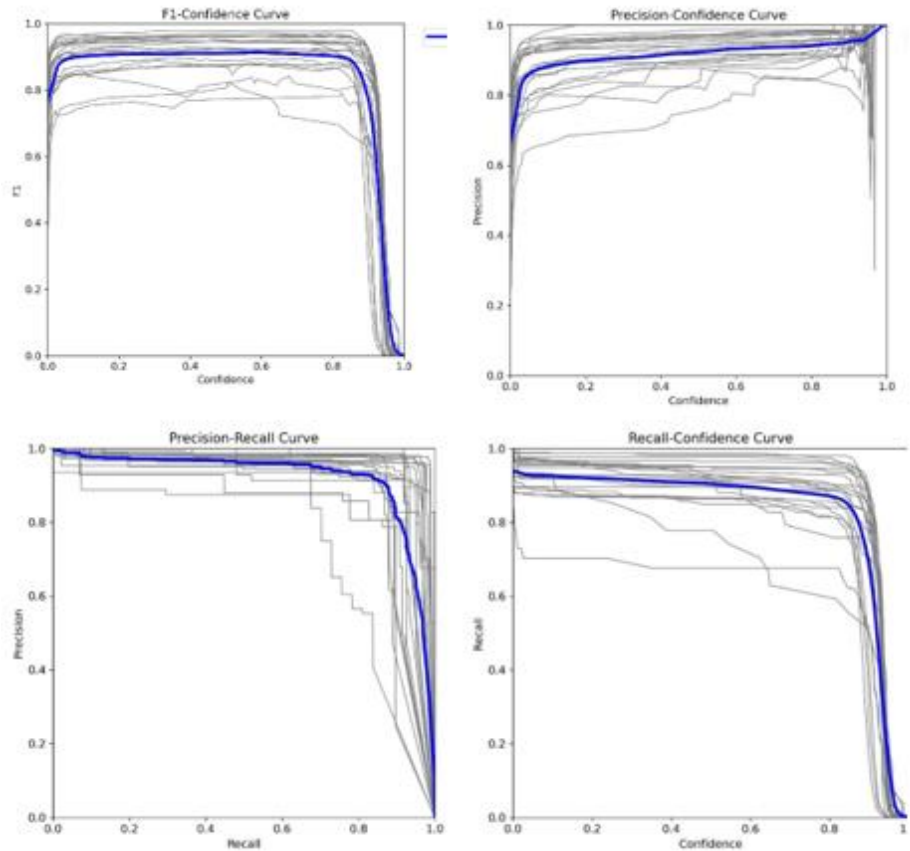

Figure 9 The confidence threshold curves

The three different kinds of losses that can occur in an object detection model are depicted in the figure (10) below. Each of the box-loss, cls-loss, and df1-loss values for the two versions are as follows: 0.4289-0.2358-0.8316 for version 2 and 0.4235-0.2315-0.8314 for version 1.

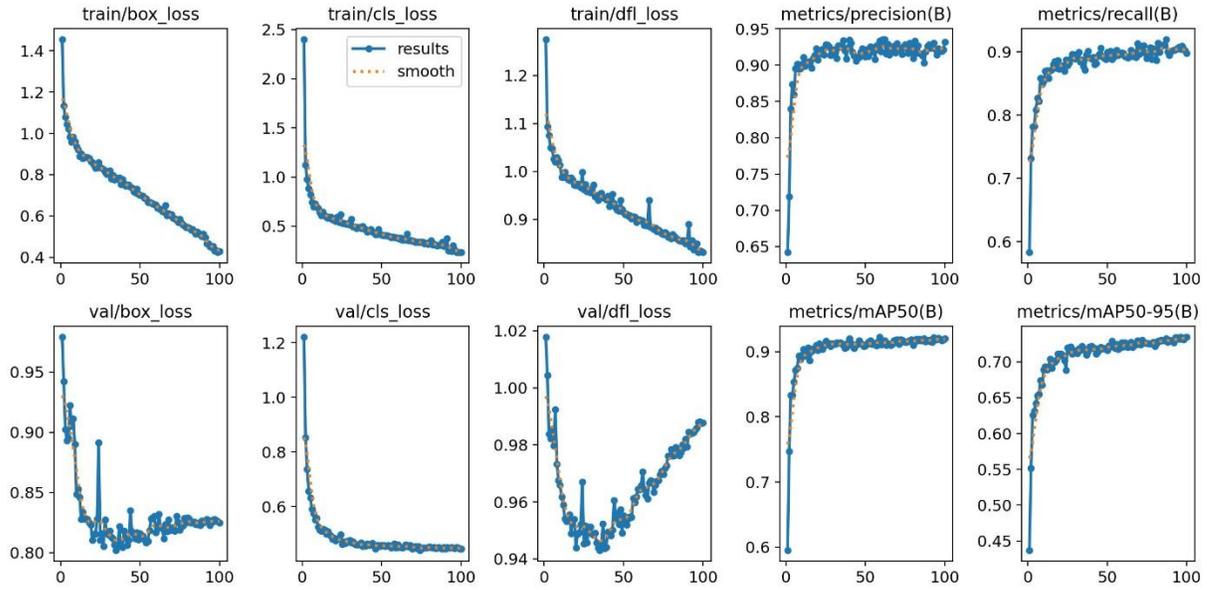

Version 1

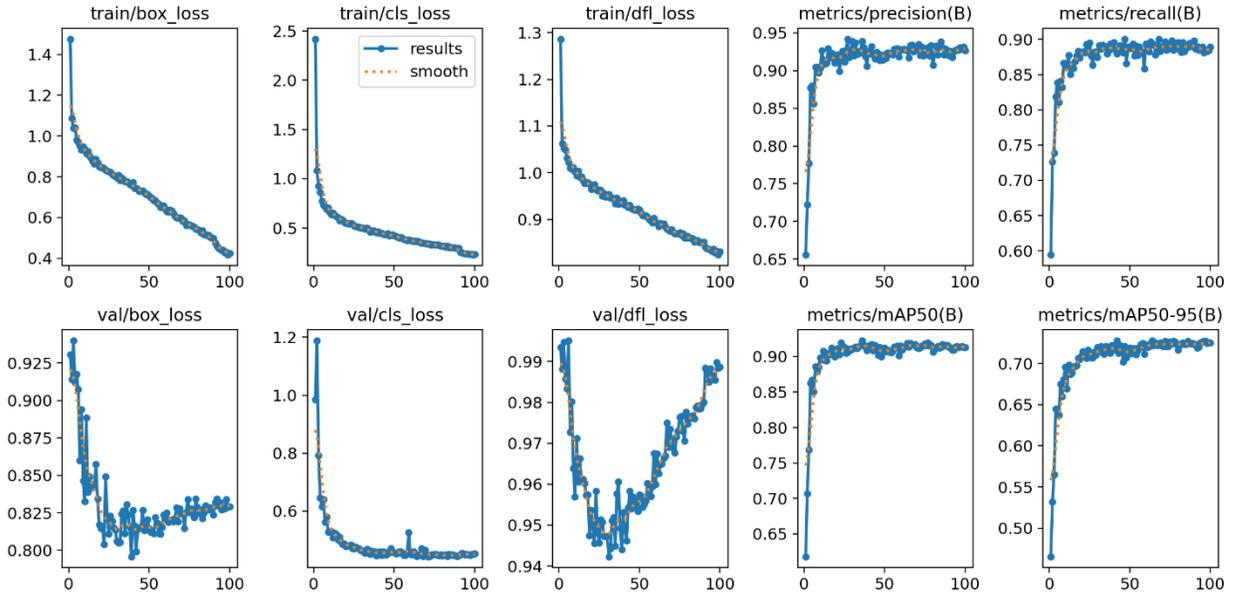

Version 2

**Figure 10** The Measurement results Curves

The classification model's performance accuracy is measured by an algorithm's top1 and top5 error rates on a classification task. The results of top1-acc and top2-acc are 96% and 100%, respectively, after 100 epochs, with a validation loss of 0.58714 and a train loss of 0.02729, as shown in *Figure 14*.

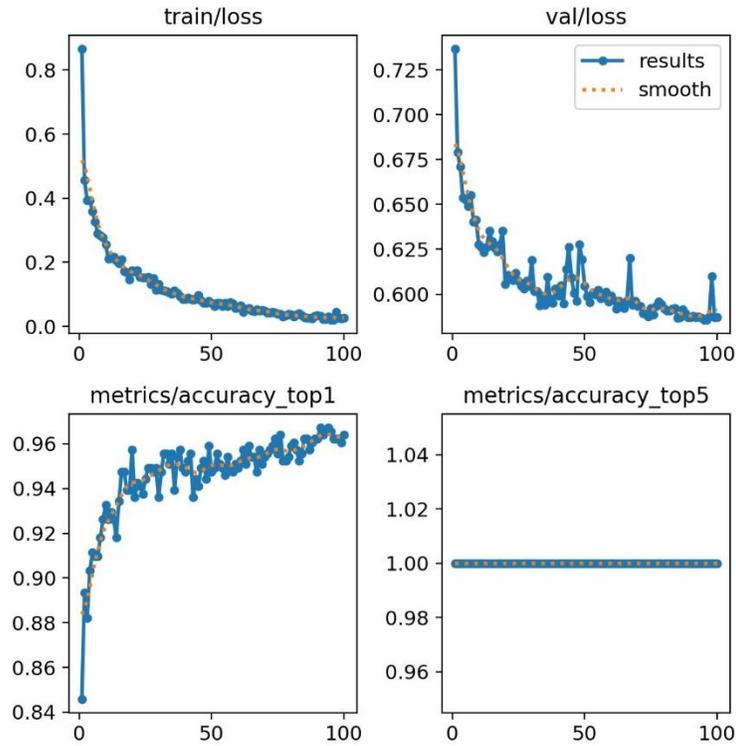

**Figure 14** The Classification Mode Accuracy

The confusion matrix for the two classes results in 0.97 for Assyrian, 0.95 for Babylonian, and 0.98 for Sumerian ratio TP as shown in *Figure 15*.

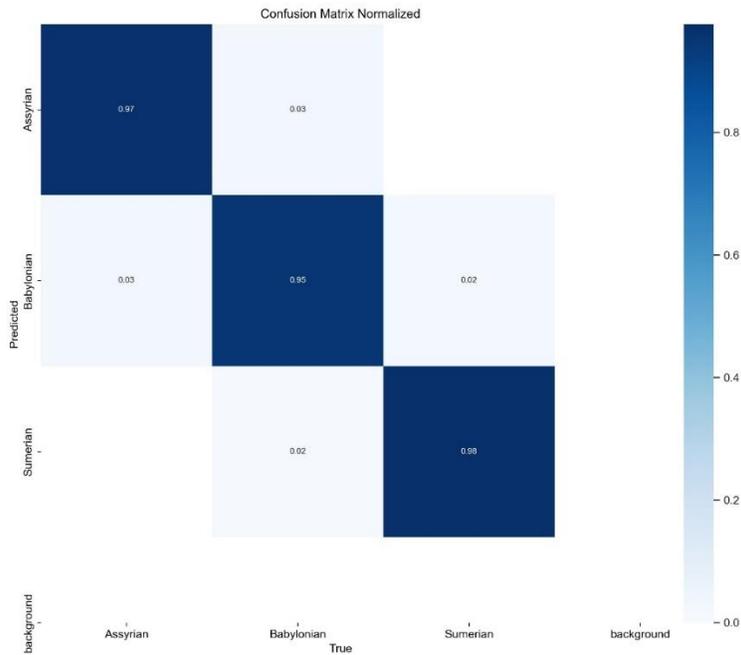

**Figure 15** Confusion Matrix of Classification Model

The generated results are regarded as being of very high quality for models. The training and testing results for classification reached the best results; however, we also need to add more photos to the dataset to expand the diversity of images and strengthen the training.

## 4. Results of two models

The forecasted outcomes for the 22 Hebrew letters indicate that each letter is encompassed by a bounding box, with the label representing the pronunciation of each letter. *Figure 16* displays the forecasted outcomes.

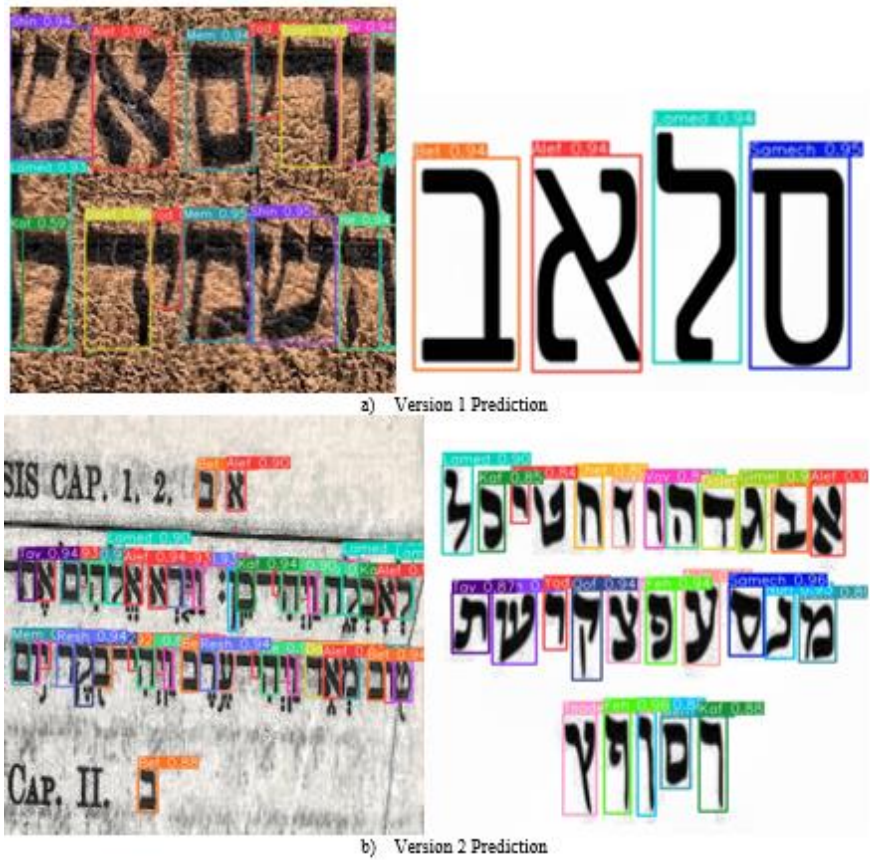

**Figure 16** Displays the Forecasted Outcomes

Figure 17 shows the classification results. The outcome (A) indicates with 100% certainty that the image is associated with the Sumerian script and 0% certainty that it is associated with the Sumerian and Babylonian script. Regarding result B, there is an 85% likelihood that the image is of Assyrian origin, 15% of Babylonian, and a 0% likelihood that it is a Sumerian script. The results (C) were 100% for Assyrian and 0% for Sumerian and Babylonian. The last result (D) was 100% Babylonian and 0% for the Sumerian and Assyrian script.

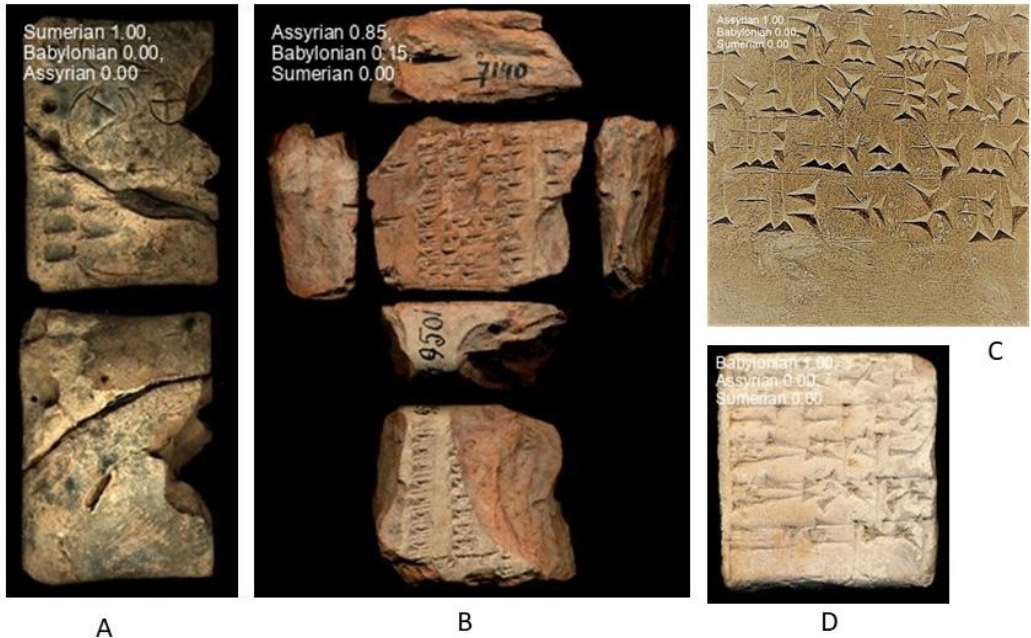

**Figure 17** The Classification Model Results

## 5. Conclusion and Discussion

Accomplishing a mAP50 score of 92%, the Hebrew characters were precisely identified. The Hebrew characters are there, as the identification procedure verified, and they match the earlier shown pictures. Low accuracy in the acquired results was caused by the difficulty in creating strong data sets due to the small number of plates at the museum. Moreover, we have achieved very good results by using the sophisticated yolov8x method, which will be improved in the next studies.

Precision is the metric used to gauge how well the model can identify real occurrences; it successfully lowers false detections by 93.2%. The Recall rate reached 89.8%, which is important to evaluate the model's precision in identifying real items. Yolov8x was also used to classify the top 5 100% and the top 1 96%.

The number of documents now available, unclear letters, damaged papers, and certain papers subjected to environmental conditions that resulted in some of them being ruined are only a few of the numerous restrictions experienced. It is hard to distinguish between the various letters—up to 472 distinct kinds of letters with similar traits. It is more complicated since Hebrew inscriptions sometimes lack gaps between letters.

Gathering clear, high-resolution information and a large number of photos is necessary to start and progress such a study. We strongly advise trying out other approaches to find any signs that could improve the results with this specific dataset.

**References:**

[1] J. Terven and D. Cordova-Esparza, "A Comprehensive Review of YOLO: From YOLOv1 and Beyond," Apr. 2023, [Online]. Available: http://arxiv.org/abs/2304.00501

[2] H. Lahoud, D. L. Share, and A. Shechter, "A developmental study of eye movements in Hebrew word reading: the effects of word familiarity, word length, and reading proficiency," *Front Psychol*, vol. 14, 2023, doi: 10.3389/fpsyg.2023.1052755.

[3] D. Fisseler, F. Weichert, G. Müller, and M. Cammarosano, *Towards an interactive and automated script feature Analysis of 3D scanned cuneiform tablets*. 2013.

[4] T. Dencker, P. Klinkisch, S. M. Maul, and B. Ommer, "Deep learning of cuneiform sign detection with weak supervision using transliteration alignment," *PLoS One*, vol. 15, no. 12 December, Dec. 2020, doi: 10.1371/journal.pone.0243039.

[5] H. Yi, B. Liu, B. Zhao, and E. Liu, "Small Object Detection Algorithm Based on Improved YOLOv8 for Remote Sensing," *IEEE J Sel Top Appl Earth Obs Remote Sens*, 2023, doi: 10.1109/JSTARS.2023.3339235.

[6] E. A. Saeed, A. D. Jasim, and M. A. Abdul Malik, "Deciphering the past: enhancing Assyrian Cuneiform recognition with YOLOv8 object detection," *International Journal of Advanced Technology and Engineering Exploration*, vol. 10, no. 109, pp. 1604–1621, Dec. 2023, doi: 10.19101/IJATEE.2023.10102331.

[7] T. L. Tobing, S. Y. Yayilgan, S. George, and T. Elgvin, "Isolated Handwritten Character Recognition of Ancient Hebrew Manuscripts," in *Archiving 2022: Expanding Connections Across Digital Cultural Heritage - Final Program and Proceedings*, Society for Imaging Science and Technology, 2022, pp. 35–39. doi: 10.2352/issn.2168-3204.2022.19.1.08.

[8] E. Stötzner, T. Homburg, J. P. Bullenkamp, and H. Mara, "R-CNN based Polygonal Wedge Detection Learned from Annotated 3D Renderings and Mapped Photographs of Open Data Cuneiform Tablets," 2023, doi: 10.2312/gch.20231157.

[9] E. Stötzner, T. Homburg, and H. Mara, "CNN based Cuneiform Sign Detection Learned from Annotated 3D Renderings and Mapped Photographs with Illumination Augmentation," in *Proceedings of the IEEE/CVF International Conference on Computer Vision*, 2023, pp. 1680–1688.

[10] L. Rothacker, D. Fisseler, G. G. W. Müller, F. Weichert, and G. A. Fink, "Retrieving Cuneiform Structures in a Segmentation-free Word Spotting Framework,"

Association for Computing Machinery (ACM), Aug. 2015, pp. 129–136. doi: 10.1145/2809544.2809562.

[11]   A. Hamplová, D. Franc, J. Pavlíček, A. Romach, and S. Gordin, "Cuneiform Reading Using Computer Vision Algorithms," in *Proceedings of the 2022 5th International Conference on Signal Processing and Machine Learning*, 2022, pp. 242–245.

[12]   "YOLOv8 Object Detection Model." Accessed: Oct. 06, 2023. [Online]. Available: https://roboflow.com/model/yolov8

[13]   Z. Li, C. Peng, G. Yu, X. Zhang, Y. Deng, and J. Sun, "DetNet: A Backbone network for Object Detection," Apr. 2018, [Online]. Available: http://arxiv.org/abs/1804.06215

[14]   X. Li *et al.*, "Generalized Focal Loss: Learning Qualified and Distributed Bounding Boxes for Dense Object Detection."

[15]   Z. Zheng, P. Wang, W. Liu, J. Li, R. Ye, and D. Ren, "Distance-IoU Loss: Faster and Better Learning for Bounding Box Regression," 2016. [Online]. Available: https://github.com/Zzh-tju/DIoU.

[16]   Y. Lee, T. Kim, and S.-Y. Lee, "Voice Imitating Text-to-Speech Neural Networks," Jun. 2018, [Online]. Available: http://arxiv.org/abs/1806.00927

[17]   J. Cho, K. Lee, E. Shin, G. Choy, and S. Do, "How much data is needed to train a medical image deep learning system to achieve necessary high accuracy?," Nov. 2015, [Online]. Available: http://arxiv.org/abs/1511.06348